\documentclass[runningheads]{llncs}
\usepackage{graphicx}
\usepackage{listings}
\usepackage{amsmath}
\usepackage{amssymb}
\usepackage[disable]{todonotes}
\usepackage{booktabs}
\usepackage{float}
\usepackage{multirow}
\usepackage{lipsum}
\usepackage{algorithm}
\usepackage{algpseudocode}
\usepackage{svg}

\DeclareMathOperator{\sel}{z}
\DeclareMathOperator{\cf}{x}

\DeclareMathOperator{\cacq}{\mathcal{A}}
\DeclareMathOperator{\csel}{\mathcal{C}}
\DeclareMathOperator*{\argmax}{\arg\!\max}

\begin{document}
\title{Active Selection of Classification Features}
\author{Thomas T. Kok$^{1,2}$\orcidID{0000-0002-4019-5629} \and
Rachel M. Brouwer$^3$ \and %
Rene M. Mandl$^3$ \and %
Hugo G. Schnack$^3$\orcidID{0000-0002-4620-3853} \and
Georg Krempl$^2$\orcidID{0000-0002-4153-2594} 
}
\authorrunning{Th. Kok et al.}
\institute{
IDLab, Ghent University - imec, Belgium \and 
Dep. of Information \& Computing Sc., Utrecht University, The Netherlands\\  %
\and
Department of Psychiatry, UMCU Brain Center, University Medical Center Utrecht, Utrecht University, Utrecht, The Netherlands  
\\ \email{thomas.kok@ugent.be}, \email{g.m.krempl@uu.nl} }
\maketitle              
\begin{abstract}
Some data analysis applications comprise datasets, where explanatory variables are expensive or tedious to acquire, but auxiliary data are readily available and might help to construct an insightful training set. An example is neuroimaging research on mental disorders, specifically learning a diagnosis/prognosis model based on variables derived from expensive Magnetic Resonance Imaging (MRI) scans, which often requires large sample sizes. Auxiliary data, such as demographics, might help in selecting a smaller sample that comprises the individuals with the most informative MRI scans. In active learning literature, this problem has not yet been studied, despite promising results in related problem settings that concern the selection of instances or instance-feature pairs.

Therefore, we formulate this complementary problem of Active Selection of Classification Features (ASCF): Given a primary task, which requires to learn a model $f:\cf \rightarrow y$ to explain/predict the relationship between an expensive-to-acquire set of variables $\cf$ and a class label $y$. Then, the ASCF-task is to use a set of readily available selection variables $\sel$ to select these instances, that will improve the primary task's performance most when acquiring their expensive features $\cf$ and including them to the primary training set.

We propose two utility-based approaches for this problem, and evaluate their performance on three public real-world benchmark datasets. In addition, we illustrate the use of these approaches to efficiently acquire MRI scans in the context of neuroimaging research on mental disorders, based on a simulated study design with real MRI data.

\end{abstract}

\section{Introduction}
In data analysis, the acquisition of data and inductive modelling are traditionally performed in distinct phases. However, combining them offers the possible advantage to focus acquisition efforts on the most insightful data, in particular when data are expensive or tedious to obtain. 
This has been acknowledged by a rich literature on selective sampling and active learning of \emph{labels} \cite{Settles2012}. In contrast, the active selection of data concerning \emph{features} has received little attention yet \cite{AttenbergEtal2011,BunseMorik2019,Kottke2016a}. 
In particular, there are data analysis applications where data on some features are not readily available and expensive to obtain, while data on other features are cheap. An example is neuroimaging research on mental disorders: Here, often large sample sizes are required, specifically when predicting diagnosis or prognosis. In current classification studies based on Magnetic Resonance Imaging (MRI), sample sizes required are several hundreds of subjects. MRI being a complex and expensive measurement tool, these numbers are hard to acquire. Given the heterogeneity in diagnoses and brain morphology, not all subjects are informative for answering the interdisciplinary research question: some subjects are 'redundant' and others are very informative. However, prior to acquiring expensive MRI data, other data might readily be available on each individual, such as their demographics or questionnaire answers. Such 'auxiliary' data might help in selecting those individuals, whose brain scans are expected to be (the most) informative. Unfortunately, there is currently no method to identify these individuals. 
This motivates the novel active learning problem of \textbf{Actively Selecting Classification Features} (ASCF) addressed in this paper: \emph{Given a primary task, which is to learn a model $f:\cf \rightarrow y$, wherein a set of expensive, yet-to-be-acquired \emph{classification} features $\cf$ is used to explain/predict a class label $y$, the ASCF task consists of using a set of readily available, cheap \emph{selection} features $\sel$ to construct the training set for the primary task. Thus, the aim is to use $\sel$ to select these instances, for which acquiring $\cf$ and adding them to the training set improves the primary task's model $f:\cf \rightarrow y$ the most. In the unsupervised variant of this problem, the labels $y$ are unknown during selection, while in the supervised variant they are readily available}

As we will show when reviewing the literature in Section \ref{sec:relwork}, this problem complements existing research in active learning, in particular on the related problems of instance completion, active feature acquisition, and active class selection. Ultimately, our aim is an ASCF-approach that guides researchers to compose the primary task's training sample in the most efficient way, thereby hopefully leading to significant reductions in sample size and costs. For this purpose, we present one supervised and one unsupervised utility-based ASCF-approach in Section \ref{sec:approach}\footnote{The code of their implementations is available as open source here: \url{https://github.com/thomastkok/active-selection-of-classification-features}.}. Following a benchmark evaluation on three real-world datasets in Section \ref{sec:exp}, we will conclude with a case study in Section \ref{sec:exp:case}, where we will apply our approach in a supervised setting to simulate the efficient 'acquisition' of MRI brain scans from a large in-house database and evaluate the resulting predictive model's performance.

\section{Related Work}\label{sec:relwork}
The active selection and acquisition of data during data modelling has been studied under the umbrella terms of active learning and selective sampling \cite{KumarGupta2020,Settles2012}. The predominantly considered problem in this literature is the active selection of instances for labelling. That is, given a pool or stream of instances with only their feature values known, the aim is to select instances to acquire their label. Some research has also been dedicated to problems related to the acquisition of features. These problems are \emph{Active Feature Acquisition} (AFA), \emph{Active Feature Selection} (AFS) and \emph{Active Class Selection} (ACS). These are closely related to the problem of Actively Selecting Classification Features considered here. However, in contrast to ASCF, they don't consider a separate set of selection features but rather consider all features for selection (of features/instances) as well as for classification. Thus, they are not directly applicable when the classification features ($x$) differ from the features ($z$) that are available during selection, as in the case of ASCF. Nevertheless, the following review of these approaches will serve as a starting point for deriving our ASCF-approaches in Section \ref{sec:approach}.

In Active Feature Acquisition \cite{Melville2004,Melville2005,SaartsechanskyMelvilleProvost2009,Zheng2006,Zheng2002}, the aim is to select and query the missing classification feature-value pairs in a dataset, which are deemed to be the most useful to improve the prediction model. That is, for which instances to acquire values of $x$.
Two alternative problem formulations exist, which differ by the number of instance values that are queried simultaneously: 
The most common one is \emph{active feature-value acquisition} \cite{SaartsechanskyMelvilleProvost2009}, where each feature-value pair is queried individually, and \emph{instance completion} \cite{Zheng2002}, where the missing features of an instance are queried all at once.

In active feature-value acquisition, as defined in  \cite{SaartsechanskyMelvilleProvost2009}, missing feature-value pairs are queried, given an incomplete feature matrix $F$, a complete label set and a cost matrix $C$ corresponding to the feature matrix. The aim is to construct the best-performing classifier, given that a query $F_{i, j}$ for the value of the $i$-th feature of the $j$-th training instance can be placed at cost $C_{i, j}$. The cost matrix $C$ is optional, with a matrix of ones as default (equal costs).
Of particular relevance for our work are the approaches based on \emph{Sampled Expected Utility} proposed in \cite{Melville2004,Melville2005}. 
These approaches first compute a score for each potential query, which indicates the (expected) increase in performance of the classifier given the corresponding feature value is acquired. Then, they select the query with the highest score, acquire the corresponding feature-values, and repeat the process after updating the classifier. 
Specifically, as defined in \cite{Melville2005}, for a discrete feature $x_i$ with $k = 1, 2, \dots K$ possible feature values, the expected utility score is
\begin{equation}
    score(x_{i,j}) = \sum_{k = 1}^{K} P(F_{i,j} = V_k) \cdot U(F_{i,j} = V_k)
\end{equation}
where $P(F_{i,j} = V_k)$ is the probability of the $i$-th feature in the $j$-th instance having the $k$-th feature value, and $U(F_{i,j} = V_k)$ is the utility of acquiring that feature value. The latter is calculated as difference in classification performance before and after incorporating the feature value in the training set, and divided by the cost of this query.
While the idea of estimating the expected utility from selecting a feature-value pair, and subsequently selecting the feature-value pair with the maximum expected utility, is at the core of most active feature-value acquisition approaches, they differ in the way they estimate the utility and the expected value distribution of the missing features.

\emph{Instance Completion}, the other variant of the Active Feature Acquisition problem, was introduction by \cite{Zheng2002}. Here, all unknown features of an instance are queried at once. It is assumed that the labels are available, while missing features occur in some instances. The goal is to build a classifier using a subset of $K$ instances with all available features, and to improve the performance when compared to using only features available in all $N$ instances. 
The approaches proposed by \cite{Zheng2006} also follow the estimation of values and utility framework. Two main approaches were proposed: Acquisition based on the Variance of Imputed Data (\emph{AVID}) and Goal Oriented Data Acquisition (\emph{GODA}). As their names indicate, the former estimates the utility on the variance of the predicted value, while the latter estimates the utility based on the change in performance measure with the predicted values.

In Active Feature Selection \cite{BilgicGetoor2007}, the problem is to select individual features, whose values are subsequently acquired for all instances at once. Thus, it complements the previously discussed instance completion, where all missing features were acquired simultaneously for one instance. 

Active Class Selection \cite{Lomasky2007} 'inverts' the role of features and labels as selection target from the conventional active learning setting: Instead of using features to select an instance among a set of candidate instances, and subsequently querying its label, the selection specifies a label, for which then a novel instance with corresponding label and complete feature vector is queried. 
Five different active class selection approaches were proposed in \cite{Lomasky2007}, the best performing one being \emph{redistricting}. In each iteration, this approach selects instances from the most volatile class. This is the class that comprises the most instances that were classified differently in the previous iteration.
More recently, approaches based on probabilistic active learning have been proposed for this problem in \cite{Kottke2016a} and \cite{BunseMorik2019}, which were shown to yield superior performance in comparison to redistricting.

Further similar approaches based on reinforcement learning and budgeted learning exist, but generally focus on an integrated training and test phase, or still have feature-related costs during the test phase \cite{HeEtal2012,ContardoEtal2016,KachueeEtal2018}.

\section{Utility-Based Active Selection of Classification Features}
\label{sec:approach}

Given are a set $\csel$ of candidate instances, each with known selection feature vector $\sel$ and (possibly already known) class label $y$, but unknown classification feature vector $\cf$, i.e., $\csel = \{ (\sel, y), \cdots \}$, and a set $\cacq$ of instances with already acquired classification features, i.e., $\cacq = \{ (\sel, \cf, y), \cdots \}$, which might be initially empty. We propose two approaches for the active selection of classification features, i.e., for selecting instances from $\csel$ for acquiring their classification feature $\cf$ and moving them to $\cacq$. Both follow the utility-principle in \cite{SaartsechanskyMelvilleProvost2009}, by estimating the utility of acquiring the classification feature vector $\cf$ of an instance.
To this end, we first formulate an auxiliary (potentially multiple multivariate) linear regression problem
$h: \sel \rightarrow \cf$, which we use to predict the most likely classification feature vector $\cf$ of each candidate instance in $\csel$. Based on this auxiliary model, we propose two approaches in the next subsections. Both differ in their data requirements and the utility measure they use:  
The first addresses a potentially \emph{unsupervised} case, i.e., it does not require the class labels $y$ to be known during selection, and uses the \emph{variance in the imputation of the missing data} as measure for utility. 
The second one is designed for \emph{supervised} settings, where the class labels are already known during selection, as it is the case in our case study. This latter approach uses the \emph{probability of misclassification} as measure for utility.

\subsection{Unsupervised, Imputation Variance-Based Variant (U-ASCF)}
Designed for cases, where the class label might not be available during instance selection, this first variant uses an ensemble of $B$ estimators $h_b: \sel_b \rightarrow \cf_b$, which are obtained by bootstrapping from the already acquired  data $\cacq$. The variance in the imputations of these bootstrapped estimators is used as an indicator of the resulting utility: The greater the variance, the higher is the imputation uncertainty and the resulting utility of acquiring the classification features of that instance. In analogy to \cite{Zheng2002}, we define this utility as follows:
\begin{equation}
    \label{eqn:u_avid}
    U(\sel, \Theta) = \frac{1}{D} \cdot \sum_{d=1}^{D}{var(\lbrace\forall \theta \in \Theta: h_\theta(\sel)_d\rbrace)}
\end{equation}
where $h_\theta(\sel)_d$ is the estimate for the $d$-th dependent variable (i.e., the $d$-th classification feature) by a regression model trained on a bootstrapped version $\theta$ of $\cacq = (\sel, \cf)$, and $\Theta$ is the set of $B$ imputation models. Thus, as utility we use the average over the variances in the $B$ boostrapped estimates of the individual classification features. As shown in the pseudocode in algorithm \ref{alg:avid}, we first train the ensemble of bootstrap estimators, use them to compute this utility for all candidate instances in $\csel$, and then acquire the classification features of the instance with the highest utility.

\begin{algorithm}
\caption{Unsupervised, Imputation-Variance Based Approach}\label{alg:avid}
\begin{algorithmic}[0]
\Procedure{U-ASCF}{$\cacq=\{(\sel_{acq\_1}, \cf_{acq\_1}),\cdots\}$, $\csel=\{(\sel_{cand\_1}),\cdots\}, B$}
    \State $\Theta \gets []$
    \For{$\theta \gets 1,2,\cdots B$}
        \State $\cacq_\theta \gets \mbox{sample}(\cacq)$
        \State $\Theta \gets \Theta \cup \{ h_\theta \gets \mbox{trainregressor}(\cacq_\theta) \}$\Comment{Train $h_\theta: \sel \rightarrow \cf$}
    \EndFor
    \For{each unacquired instance $j \in \csel$}
        \State $\mbox{feature\_estimates}_j \gets \forall \theta \in \Theta: h_{\theta}(\sel_j)$
        \State $U_j \gets  avg(var(\mbox{feature\_estimates}_j))$ \Comment{Use Eqn. \ref{eqn:u_avid}}
    \EndFor
    \State $j^* = \argmax_j U_j$ \Comment{Select instance $j^*$}
    \State $\cf_{j^*} \gets acquire(j^*)$  \Comment{Acquire class. feat.}
    \State \textbf{return} $\cacq' = \cacq \cup \{\sel_{j^*}, \cf_{j^*}\}, \csel' = \csel \setminus \{\sel_{j^*}\}$
\EndProcedure
\end{algorithmic}
\end{algorithm}

\subsection{Supervised, Probabilistic Selection Variant (S-ASCF) }
For cases where the class labels are already available during selection, and probabilistic classifiers are used in the primary classification task $f: \cf \rightarrow y$, their class' posterior estimates allow to adapt the probability-based utility measure discussed in \cite{Dhurandhar2015} to our problem setting. In \cite{Dhurandhar2015}, for a given instance with unknown feature $x$ this measure is defined as 
\begin{equation}
    score_x = \frac{p (1 - p)}{(-2b + 1)p + b^2}
\end{equation}
where $p$ corresponds to the probability of misclassification, 
and $b \in [0,1]$ is an asymmetry parameter with default value
$0.5 + \frac{1}{2 \cdot |\cacq|}$ for $|\cacq|$ corresponding to the number of already previously queried and completed instances, such that $b$ converges towards $0.5$ as the number of queried instances increases.

As shown in the pseudocode in algorithm \ref{alg:prob}, we first train the primary classifier  ${f}: \cf \rightarrow y$ and the auxiliary regressor  ${h}: \sel \rightarrow \cf$ on the set of already acquired instances $\cacq$. 
For the candidate instances in $\csel$, we then use their selection features $\sel$ and the auxiliary regression model to obtain estimates for their classification features $\hat{\cf} = h(\sel)$, on which we deploy the primary classifier to get posterior estimates $f(\cf)_{prob}$ of their most likely class. This is then used in a supervised probability-based utility measure adopted to the ASCF-task:
\begin{equation}
    \label{eqn:u_prob}
    U(f, h, \sel, y) = \frac{P({f}(h(\sel)) \neq y) \cdot (1 - P({f}(h(\sel)) \neq y))}{(-2b + 1) \cdot P({f}(h(\sel)) \neq y) + b^2}
\end{equation}

\begin{algorithm}
\caption{Supervised, Probability-Based ASCF}\label{alg:prob}
\begin{algorithmic}[0]
\Procedure{S-ASCF}{$\cacq=\{(\sel_{acq\_1}, \cf_{acq\_1}, y_{acq\_1}),\cdots\}$, $\csel=\{(\sel_{cand\_1},y_{cand\_1}),\cdots \}$}
    \State $f \gets \mbox{trainprobclassifier}(\cacq)$ \Comment{train $f: \cf \rightarrow y$}
    \State $h \gets \mbox{trainregressor}(\cacq)$\Comment{train $h: \sel \rightarrow \cf$}
    \State $b \gets \frac{1}{2} + \frac{1}{2 * \lvert \cacq \rvert}$ \Comment{Estimate metaparameter}
    \For{each unacquired instance $j \in \csel$}
        \State $\cf_j \gets h(\sel_j)$
        \State $p \gets 1 - f(\cf_j)_{prob}$
        \State $U_j \gets \frac{p (1 - p)}{(-2b + 1)p + b^2}$ \Comment{Use Eq. \ref{eqn:u_prob}}
    \EndFor
    \State $j^* = \argmax_j U_j$ \Comment{Select instance $j^*$}
    \State $\cf_{j^*} \gets acquire({j^*})$  \Comment{Acquire class. feat.}
    \State \textbf{return} $\cacq' = \cacq \cup \{\sel_{j^*}, \cf_{j^*}, y_{j^*}\}$,  $\csel' = \csel \setminus \{\sel_{j^*}, y_{j^*}\}$
\EndProcedure
\end{algorithmic}
\end{algorithm}

\section{Experimental Results}\label{sec:exp}
We have designed a series of experiments, which simulate a real-world deployment of each ACFS-approach (U-ASCF, S-ASCF, and random selection as baseline).
Therein, all selection feature values $\sel$ as well as the class labels $y$ are readily available\footnote{The unsupervised approach U-ASCF does not use these labels during selection.} for each instance of the training set (which corresponds to $\csel$), while all classification feature values $\cf$ are initially concealed (and have to be acquired by the selection approach, thus $\cacq = \emptyset$ initially). The primary classifier is trained on the selected subset of the training set by the relevant approach, and tested on the classification feature values and corresponding class labels of the test set. The predictions of this primary classifier are then compared to the true class labels, and evaluated.

All approaches use the same splits into training and test sets, obtained via repeating 5-fold cross-validation 10 times, and the same classifier technique with a priori tuned hyperparameters. For the experiments, $B$ is set to 10.  As classifier, \emph{logistic regression} \cite{Berkson1944} with $C = 1.0 \text{ and } L2$ regularization is used\footnote{Experiments with SVMs were also performed with similar results, see our website.}.
For each selection step, the next instance to acquire is selected, and the training set and primary classifier are updated, and its resulting performance on the test set is evaluated and plotted  in a comparison with that of its competitors.

\subsubsection{Real-World Benchmark Datasets}\label{sec:exp:data} 
For reproducibility of the comparative evaluation, three benchmark datasets from the public UCI Machine Learning Repository
\cite{DuaGraff2019} are used. The datasets should be binary classification
datasets, with a relatively large number of numerical features.
Since these datasets do not provide distinguished selection and classification features, we manually split the features in these datasets into two groups, such that the simpler and easier obtainable ones were assigned to the selection set. The active selection of instances for the training set does \emph{not} alter instances that are in the hold out test set.
We have used the following datasets and selection features (all remaining features are classification features):

\begin{enumerate}
    \item \textit{Breast Cancer Coimbra} with Age and BMI as selection features 
    \item \textit{Heart Disease} with Age, Sex and Chest Pain Type as selection features
    \item \textit{Wine} with Alcohol, Color intensity and Hue as selection features
\end{enumerate}

\subsection{Comparative Results}\label{sec:exp:res}

\begin{figure}[h]
    \centering
    \includegraphics[width=\textwidth]{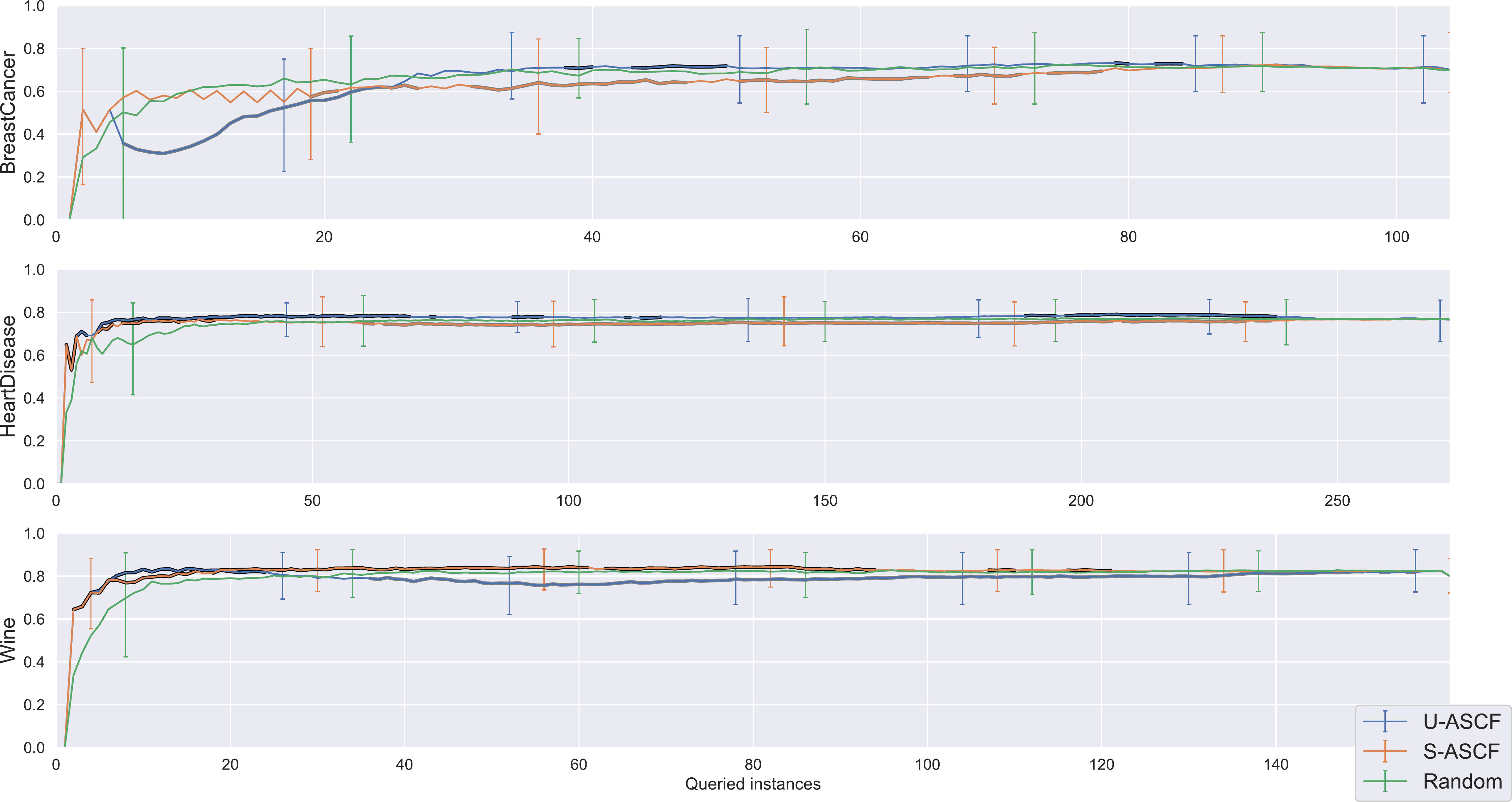}
    %\includesvg[width=\textwidth]{bm_plot.svg}
    \caption{Comparison of the F1-score of the two utility methods and the
    random baseline, on the UCI datasets. Error bars denote the 10th and 90th percentile of the results. Black outlines denote statistically significant difference compared to random selection, with $\alpha = 0.1$.}
    \label{fig:res_benchmark}
\end{figure}

The results in terms of the F1-score, shown on ordinate, defined as harmonic mean of precision and recall, for both proposed approaches and a random baseline are shown in Figure \ref{fig:res_benchmark}. The learning steps shown on the abscissa range from 0 to all instances having been acquired for their classification feature. The 10th and 90th percentile are displayed as error bars on the learning curves, slightly spread out for each approach as to not overlap. The objective is fast convergence to high F1 scores, while with an increasing number of queried instances -and thus converging training sets- all approaches should converge to the same performance.

Using a one-sided Wilcoxon signed-ranked test with $\alpha = 0.1$, for each step of the learning curve and each approach, we test the statistical significance in the F1-score difference compared to the random selection baseline. All steps with such a significant difference in performance compared to random sampling are then highlighted with a outlines in black.

The results are differing, depending on the
dataset (but independent of the classifier model).
For the Heart Disease dataset, the \emph{unsupervised ASCF} method performed consistently better than the random sampling baseline, both at the early segment of the learning curve as well as near the later stages. Its average performance is better at all stages of the learning curve, but differs not significantly at most points in the middle segment.  
For the Breast Cancer and Wine datasets, the relative performance varies with the learning stage: On both datasets, there is a steep learning curve initially, followed by a subsequent loss in relative comparison. On the Breast Cancer dataset, this loss is significant between 5-30 acquisitions, while U-ASCF and random are en par afterwards. For the Wine dataset, U-ASCF yields significant improvement until 25 acquisitions, but falls significantly below random later on.
The \emph{supervised, probability-based ASCF} method provides initially better performance then random. On the Wine dataset, its performance remains significantly better. On the Heart Disease dataset, the initially significantly better performance later declines to a significantly lower performance after 60 acquisitions. On the Breast Cancer dataset, this happens already after about 17 acquisitions. 
Overall, for both approaches the results vary between datasets and learning step. They are promising in early learning stages, but their relative performance does not remain consistently superior in later stages. 
In conclusion, if learning with a very limited budget is the objective and performance with few acquisition matters, they showed in all but the Breast Cancer dataset an improvement over random sampling.

\subsection{Case Study}\label{sec:exp:case}
For this problem setting, we considered the real-world problem of the prediction of the schizophrenia diagnosis based on MRI brain scans. Using the database of processed MRI brain images ($N > 1000$) at the Department of Psychiatry, UMC Utrecht, Netherlands, we simulated a study to answer the following neuroimaging research question: Can we classify schizophrenia patients and healthy subjects, based on their MRI brain scans? It has been previously shown that this can be done with reasonable accuracy, using several hundreds of subjects \cite{Nieuwenhuis2012}.

Using the FreeSurfer processing software \cite{Fischl2004}, the following brain features were extracted: thickness, surface area and volume of 68 cortical regions of interest, and volumes of the subcortical structures. These features were used to train diagnostic classification models. After selecting subjects with available 'cheap' selection features (age, sex, IQ), the sample available for our experiment included 633 (age, sex and IQ) or 814 (age and sex) subjects, labeled as schizophrenia patient or healthy control subject.

The MRI database forms 'the population' (sociodemographic and clinical data are available). The default approach to answer this neuroimaging research question is to include a large number ($N =$ several hundreds) of scans from 'the population', on which a classification model is learned passively. The results of this approach serve as a reference. The alternative approach is to acquire a significantly smaller number of scans by searching and selecting subjects from 'the training population'. Whether or not a new subject will be  selected to obtain an MRI scan will be based on the selection features. The performance of the active learning approach were performed in terms of classification
accuracy and reduction of sample size needed to obtain this result.

For the simulated case study, we used recursive feature elimination \cite{Guyon2002} to reduce the number of features and improve performance. It aims to reduce the number of features used recursively, using cross-validation and the logistic regression classifier model to select the least relevant features. When performing this feature elimination process\todo{if space, explain why and how (R4)}, we noted the optimal number of features to used for both instantiations of the dataset: 33 for the instances with IQ score, and 9 for all instances.

\begin{figure}[h]
    \centering
    \includegraphics[width=\textwidth]{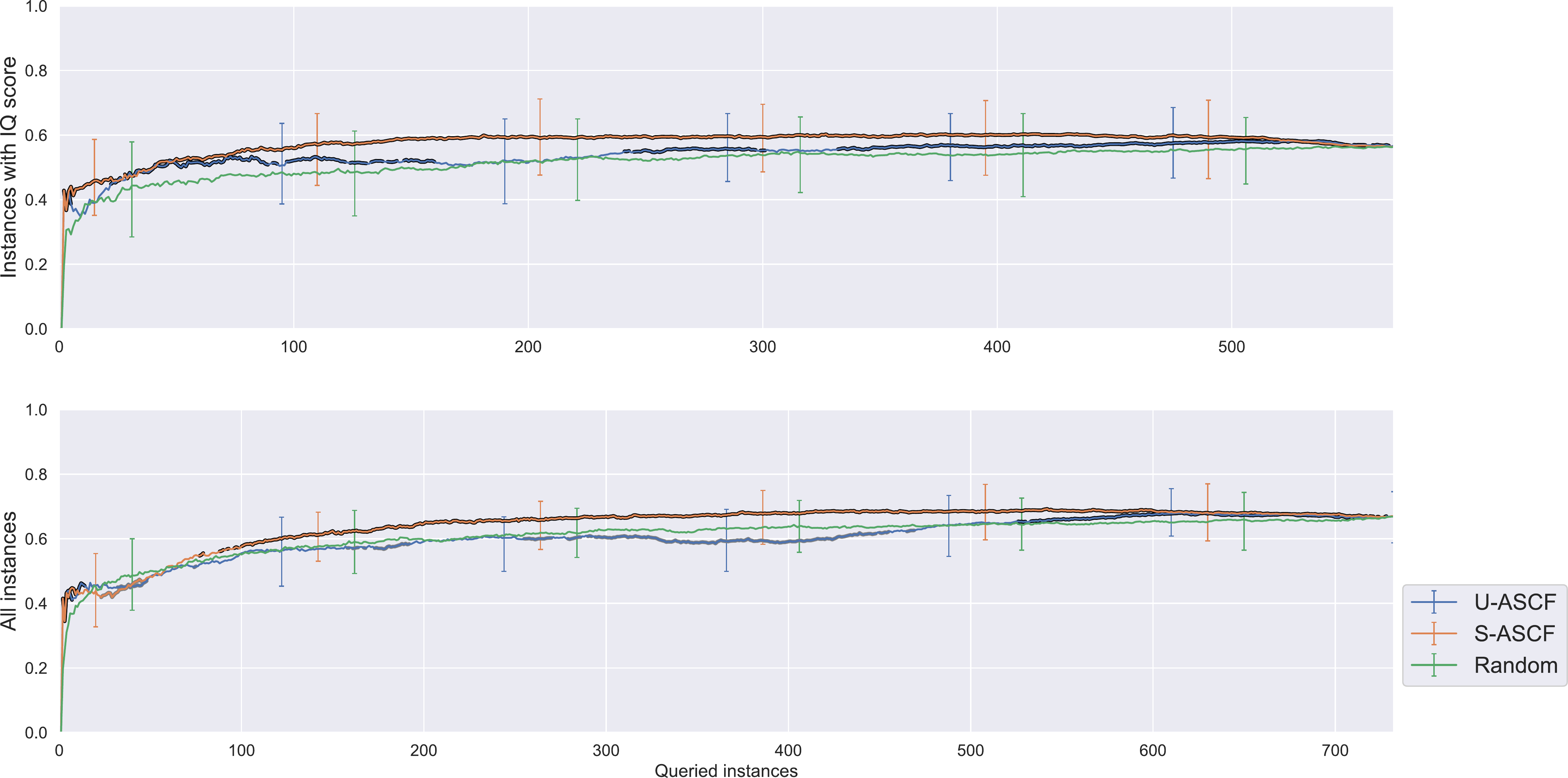}
    %\includesvg[width=\textwidth]{cs_plot.svg}
    \caption{Comparison of the F1-score of the two utility methods and the
    random baseline, on the two configurations of the case study dataset. Error bars denote the 10th and 90th percentile of the results. Black outlines denote statistically significant difference compared to random selection, with $\alpha = 0.1$.}
    \label{fig:cs_f1}
\end{figure}

As our goal for this case study is to reduce the number of instances needed with  similar performance, we will evaluate primarily using the learning curve as seen in Figure \ref{fig:cs_f1} with a focus on the mid-to-latter part of the curve.
In both configurations of the dataset, the approach using the probability-based utility method provided a consistent improvement over the random baseline. This suggests that if these approaches would have been used in the real-world scenario, either (a) the number of needed instances could have been reduced, or (b) for the same number of instances a better performance could have been obtained. This is especially supported by looking at the end of the learning curve. In the case of the learning curve of the random baseline, it consistently increases until it reaches its best performance score at the end---as makes sense, as more information is obtained. However, for our approach, the learning curve is able to reach its best performance much earlier, and even decreases
near the end (where the approach is forced to acquire also the uninformative instances). This means that these approaches are able to \emph{filter out
the instances with negative usefulness}. These are instances that do not make
sense with the rest of the dataset or the model, and thus we can say that these
approaches are able to select instances in a useful manner.

The unsupervised ASCF method is able to improve upon the random baseline in one configuration with statistical significance for a majority of the learning curve, and for the other configuration is able to statistically significantly improve upon this baseline given a higher number of sampled instances.

The supervised ASCF method provides even better results, providing a statistically significant improvement on the F1-scored over the random baseline at almost every point of the experiment for both datasets.

All in all, we can consider these approaches to be very useful in this experimentation setting, if we are to consider the F1-score performance measure. This allows for consistent and significant improvement upon the random sampling baseline.  The most consistent seem to be the supervised, probability-based ASCF approach. 
In practice, the implications of the improved F1-scoring in comparison with the implementation costs of an ASCF-protocol have to be considered. Furthermore, by focusing selection on the most informative instances, an active learning approach will introduce a bias, which is beneficial for learning the classifier in the primary task, but needs to be considered when using the obtained data for different purposes.

\section{Conclusion}\label{sec:conc}
In this paper, we have introduced a new kind of active learning problem, the Active Selection of Classification Features (ASCF): 
Given a primary task, which requires to learn a model $f:\cf \rightarrow y$ to explain/predict the relationship between a set classification features $\cf$, which are expensive to acquire, and a class label $y$. Then, the ASCF-task is to use a set of readily available, cheap selection variables $\sel$ to select these instances, that will improve the primary task's performance most when acquiring their expensive features $\cf$ and including them to the primary training set.

For this problem, we have proposed two utility-based approaches: First, an unsupervised approach that does not require labels during selection and uses the variance within data imputations as indicator of utility. Second, a supervised approach, which uses a measure adopted from probability-based active feature acquisition to select when labels in addition to selection features are available. Evaluation on three public benchmark datasets shows promising performance in initial learning stages, but also the need to improve robustness in later learning stages. A case study in the context of neuroimaging research on mental disorders indicates that the approaches are capable of reducing the number of MRI scans, that need to be acquired for reaching the same predictive performance, in comparison to random-based selection.

\subsection*{Acknowledgements}
We would like to thank Ad Feelders for valuable discussions on this topic. Furthermore, we would like to thank the SIG Applied Data Science at UU/UMCU for funding the research project "Using active learning to reduce the costs of population-based neuroimaging studies".

\bibliography{main}
\bibliographystyle{plain}
\end{document}